\documentclass[10pt,conference,compsocconf]{IEEEtran}

\usepackage{amsmath,amssymb,amsfonts}
\usepackage{algorithmic}
\usepackage{graphicx}
\usepackage{textcomp}
\usepackage{xcolor}
\def\BibTeX{{\rm B\kern-.05em{\sc i\kern-.025em b}\kern-.08em
    T\kern-.1667em\lower.7ex\hbox{E}\kern-.125emX}}

\usepackage[maxcitenames=1,mincitenames=1,backend=biber,style=ieee,natbib=true]{biblatex}
\usepackage{bm}
\usepackage{booktabs}
\usepackage{url}
\usepackage[subrefformat=parens]{subcaption}
\usepackage{algorithm}
\usepackage{algorithmic}

\newcommand{\tr}{\mathsf{T}}
\newcommand{\trace}{\operatorname{Tr}}

\begin{document}

\title{Shapley Values of Reconstruction Errors of PCA\\for Explaining Anomaly Detection}

\author{\IEEEauthorblockN{Naoya Takeishi}
\IEEEauthorblockA{\textit{Center for Advanced Intelligence Project} \\
\textit{RIKEN}\\
Tokyo, Japan \\
\url{naoya.takeishi@riken.jp}}
}

\maketitle

\begin{abstract}
We present a method to compute the Shapley values of reconstruction errors of principal component analysis (PCA), which is particularly useful in explaining the results of anomaly detection based on PCA. Because features are usually correlated when PCA-based anomaly detection is applied, care must be taken in computing a value function for the Shapley values. We utilize the probabilistic view of PCA, particularly its conditional distribution, to exactly compute a value function for the Shapely values. We also present numerical examples, which imply that the Shapley values are advantageous for explaining detected anomalies than raw reconstruction errors of each feature.
\end{abstract}

\begin{IEEEkeywords}
anomaly detection,
fault detection and isolation,
interpretable machine learning,
Shapley values,
principal component analysis
\end{IEEEkeywords}


\section{Introduction}

Anomaly detection based on machine learning has been actively studied and now plays an important role in various industrial applications such as fraud detection in finance \citep{ahmed_survey_2016}, intrusion detection \citep{garcia-teodoro_anomaly-based_2009}, and fault detection of mechanical systems \citep{khan_review_2018}.
Up to date, there have been proposed many types of anomaly detection algorithms based on different assumptions and technical principles (see, e.g., \citep{chandola_anomaly_2009,campos_evaluation_2016,goldstein_comparative_2016}).

In this work, we focus on the so-called \emph{unsupervised anomaly detection} \citep{chandola_anomaly_2009} problems, wherein the reference data used for learning a model contain only nominal, non-anomalous states of a target system.
Moreover, we concentrate on the use of \emph{reconstruction errors} as the measure of deviation from the nominal status.
In fact, unsupervised anomaly detection using reconstruction errors is a popular technical setting.
For example, reconstruction errors of principal component analysis (PCA) and autoencoders (and their variants) have been utilized in various applications (see, e.g., \citep{lakhina_diagnosing_2004,yairi_data-driven_2017,zhou_anomaly_2017,zong_deep_2018} and the references therein).

Reconstruction errors are used for not only detection but also explaining anomalies, to some extent, in practice.
Now, let us suppose that we are monitoring the reconstruction errors of data comprising multiple features.
Let $\bm{x}=[x_1 \ \cdots \ x_d]^\tr\in\mathbb{R}^d$ be a data instance (with $d$ features) and $\hat{\bm{x}}$ be a reconstruction of $\bm{x}$.
Then, the reconstruction error $e(\bm{x})$ is defined by
\begin{equation}
    e(\bm{x}) = \Vert \hat{\bm{x}} - \bm{x} \Vert_2^2 = \sum_{i=1}^d (\hat{x}_i - x_i)^2.
\label{eq:recerr_general}\end{equation}
As $e(\bm{x})$ is the sum of the reconstruction errors of the features, $x_1,\dots,x_d$, if all the features are scaled appropriately, we may anticipate that a feature having larger reconstruction errors is more suspicious.
However, is it truly explaining the features' contribution to the total reconstruction error, $e(\bm{x})$?
Not necessarily so it would be.
A large error on a feature, $x_i$, may stem from anomalous behavior of another feature, $x_j$.
Thus, by simply looking at errors of each feature, we may fail to localize detected anomalies.

This preliminary work focused an alternative mean to explain reconstruction errors for localizing detected anomalies.
In particular, we introduce a way to compute the \emph{Shapley values} \citep{shapley_value_1953} of reconstruction errors.
The Shapley value is a notion originally used to distribute a gain of a coalitional game to the players.
It has been utilized also for explaining outputs of machine learning models \citep{lipovetsky_entropy_2006,strumbelj_explaining_2014,lundberg_unified_2017}.
The Shapely value would be useful for localizing detected anomalies, too.
In an analogy to a game, the raw reconstruction error of each feature is (a part of) the final gain of a game possessed separately by each player (i.e., feature) and not necessarily represents the true contribution of each player; e.g., the gain one possesses may be generated by another.
In contrast, the Shapley value is known as a quantity that has desirable properties to measure the contribution generated by each player.

In this paper, we show how a value function for the Shapley values of reconstruction errors, particularly for PCA, is exactly computed using the probabilistic view of PCA.
Moreover, we provide numerical examples, which exhibit the potential utility of the Shapley values of reconstruction errors to explain anomaly detection.


\section{Background}

\subsection{Anomaly Detection Using Reconstruction Errors}

One of the most popular kinds of unsupervised anomaly detection methods utilizes the \emph{reconstruction errors}, which are the deviation between test data and the corresponding reconstructions computed by a model trained on a reference dataset.
Below we summarize the abstract procedures of such methods and examples of popular methodologies.

Suppose that a dataset is a set of column vectors $\bm{x} \in \mathbb{R}^d$.
As the first step of anomaly detection, we learn a model $(\bm{f}, \bm{g})$ on a dataset that contains no (or few) anomalies.
Here, $\bm{f} \colon \mathbb{R}^d \to \mathbb{R}^p$ (usually $d > p$) is a map (so-called encoder) from the original data space to some latent compressed representation space, and $\bm{g} \colon \mathbb{R}^p \to \mathbb{R}^d$ is a map (decoder) that approximately restores the original data from the compressed representations.
After learning $\bm{f}$ and $\bm{g}$, we compute the reconstruction error\footnote{We may define the reconstruction error $e(\bm{x})$ as the norm of \emph{a part of} $\hat{\bm{x}}-\bm{x}$ when the features of interest are limited. The methods discussed in this paper apply even to this case only with a slight modification.} on a new data-point $\bm{x}$ using Eq.~\eqref{eq:recerr_general} with $\hat{\bm{x}} = \bm{g}( \bm{f}(\bm{x}) )$.
If $e(\bm{x})$ is small, it is likely that the new data-point $\bm{x}$ does not deviate from the training data, and thus it is regarded as normal.
In contrast, if $e(\bm{x})$ is large, we will suspect that some features of the new data-point $\bm{x}$ are anomalous.

Among many choices of the model, $(\bm{f}, \bm{g})$, a linear model or its mixtures are popular because of their simplicity, computational efficiency, and stability.
Particularly, PCA \citep[see, e.g.,][]{PCABook_Jolliffe} is a well-known classical method for computing reconstruction errors.
In this work, we focus on the use of PCA for anomaly detection, so we review PCA in the next subsection.
Of course, a linear model may be insufficient to capture characteristics of data in practice.
In such cases, a popular alternative is to use autoencoders, which are a neural network model that can compress and approximately restore data.
For the use of autoencoders or their variants for anomaly detection, see, e.g., \citet{zhou_anomaly_2017} and \citet{zong_deep_2018}.
Finally, we note that there is a recent study by \citet{antwarg_explaining_2019} in which they try to explain the reconstruction of autoencoders using a method related to the one used in our work.
We discuss the difference between their work and ours in Section~\ref{discussion:related}.

\subsection{Principal Component Analysis}

In this work, we focus on the anomaly detection using the reconstruction errors computed via PCA \citep[see, e.g.,][]{PCABook_Jolliffe}, in which both a encoder and a decoder are linear models.
For the convenience in the main part of this work, we adopt the probabilistic view of PCA \citep{tipping_probabilistic_1999}.
In this view, the data $\bm{x}\in\mathbb{R}^d$ and the compressed representations (i.e., principal component scores) $\bm{z}\in\mathbb{R}^p$ are regarded as Gaussian random variables, and probabilistic distributions on them are defined as follows (note that \eqref{eq:ppca3} is just a direct consequence of \eqref{eq:ppca1} and \eqref{eq:ppca2}):
\begin{align}
    p(\bm{x} \mid \bm{z}) &= \mathcal{N}_{\bm{x}} ( \bm{W} \bm{z} + \bm{b}, \ \sigma^2 \bm{I} ),
    \label{eq:ppca1}
    \\
    p(\bm{z}) &= \mathcal{N}_{\bm{z}} ( \bm{0}, \ \bm{I} ),
    \label{eq:ppca2}
    \\
    p(\bm{x}) &= \mathcal{N}_{\bm{x}} ( \bm{b}, \ \bm{C} ),
    \label{eq:ppca3}
\end{align}
where $\{\bm{W},\bm{b},\sigma^2\}$ is the set of the parameters, and
\begin{equation}
    \bm{C} = \sigma^2 \bm{I} + \bm{W} \bm{W}^\tr.
\label{eq:Cmat}
\end{equation}
Here, the column of $\bm{W}\in\mathbb{R}^{d \times p}$ are principal components, $\bm{b}\in\mathbb{R}^d$ is the interception parameter, and $\sigma^2\in\mathbb{R}_{>0}$ is the observation noise variance.
It is known \citep{tipping_probabilistic_1999} that the maximum likelihood estimation of $\bm{W}$ coincides with the result of the classical (non-probabilistic) PCA in the limit of $\sigma^2 \to 0$.
The maximum likelihood estimation of $\bm{W}$ can be obtained via eigendecomposition of a data covariance matrix or iterations of the EM algorithm.

The optimal least-squares reconstruction by the probabilistic PCA \citep{tipping_probabilistic_1999} is given as
\begin{equation}
    \hat{\bm{x}} = \bm{B} \bm{x},
    \quad\text{where}\quad
    \bm{B} = \bm{W} (\bm{W}^\tr \bm{W})^{-1} \bm{W}^\tr.
\label{eq:Bmat}
\end{equation}
Therefore, the corresponding reconstruction error is given in a quadratic form:
\begin{equation}
    e(\bm{x}) = \Vert \hat{\bm{x}} - \bm{x} \Vert_2^2 = \trace \big( (\bm{I} - \bm{B}) \bm{x} \bm{x}^\tr \big).
\end{equation}

\subsection{Shapley Values}

The \emph{Shapley value} \citep{shapley_value_1953} is a concept in coalitional games.
It is a method to distribute the total gain of a game to the players and has desirable properties such as efficiency, symmetry, and linearity.
It has also been utilized in explaining machine learning models in several contexts of machine learning \citep{lipovetsky_analysis_2001,lipovetsky_entropy_2006,strumbelj_explaining_2009,strumbelj_efficient_2010,strumbelj_explaining_2014,lundberg_unified_2017}.
In these contexts, a gain is the output of a machine learning model, and a player is each of the features fed into the model.

Let $D=\{1, \dots, d\}$ be the set of feature indices, $S \subseteq D$ be an index subset, and $v(S)$ be a value function that expresses the output of a machine learning model given only the features indexed in $S$.
Then, the Shapley value of $v$ with regard to the $i$-th feature, $\varphi_i(v)$, is defined as
\begin{equation}\begin{gathered}
    \varphi_i(v) = \sum_{S \subseteq D \backslash \{i\}} \gamma(S) \big( v(S \cup \{i\}) - v(S) \big),
    \\
    \gamma(S) = \frac{\vert S \vert! \, (d - 1 - \vert S \vert)!}{d!}.
\end{gathered}\label{eq:shapley1}\end{equation}
In fact, the following expression is equivalent to \eqref{eq:shapley1}, i.e.,
\begin{equation}
    \varphi_i(v) = \frac1{d!} \sum_{O \in \pi(d)} \Big( v \big( \text{Pre}_i(O) \cup \{i\} \big) - v \big( \text{Pre}_i(O) \big) \Big),
\label{eq:shapley2}
\end{equation}
where $\pi(d)$ is the set of permutations of $(1,\dots,d)$, and $\text{Pre}_i(O)$ is the set of feature indices that precede $i$ in an order $O$; e.g., $\text{Pre}_2((1,4,2,3)) = \{1,4\}$.

There have been proposed several ways to define the value function $v(\cdot)$ to use Shapley values for explaining machine learning models.
In Shapley value regression \citep{lipovetsky_analysis_2001,lipovetsky_entropy_2006}, $v(S)$ is defined as the coefficient of determination of models using features in $S$, with which they measure contributions of features to the explained variance.
In another line of studies \citep{strumbelj_explaining_2009,strumbelj_efficient_2010,strumbelj_explaining_2014}, $v(S)$ is defined as the output of a model trained using the features in $S$.
As a na\"ive implementation of this value needs retraining of the model for different $S$ \citep{strumbelj_explaining_2009}, the consequent studies \citep{strumbelj_efficient_2010,strumbelj_explaining_2014} approximate it by integrating a full model's outputs with regard to the features absent in $S$.
A similar idea is adopted in the method called Shapley additive explanation (SHAP) \citep{lundberg_unified_2017}, which defines $v(S)$ as the output of a local surrogate model.

The exact computation of $\varphi(v)$ using \eqref{eq:shapley1} or \eqref{eq:shapley2} is intractable for a moderate number of features, $d$, because we have to consider all subsets of $D \backslash \{i\}$ for the summation, whose number increases exponentially with regard to $d$.
Hence, we need some approximation, and a couple of methods to this end are known.
A straightforward option is to approximate the summation in \eqref{eq:shapley2} simply by a Monte Carlo method, i.e., by randomly choosing \emph{some} subsets of $D \backslash \{i\}$ for the summation.
In this paper, we adopt this Monte Carlo strategy.
Another popular approximation method is the one used in SHAP \citep{lundberg_unified_2017}, namely kernel SHAP.
Kernel SHAP utilizes a sample weighting function (a kernel) with which a linear local surrogate model coincides with the approximation of the Shapley values.
Considering the latter type of approximation is an open problem in our context.


\section{Shapley Values of Reconstruction Errors\\of PCA}

In this work, we aim to explain the reconstruction errors of PCA using the idea of the Shapley values.
In other words, we would like to compute the Shapley values when the value function, $v(S)$, is defined via the reconstruction errors of PCA on the features indexed in $S$.
To this end, the main problem of ours is how to define $v(S)$ in a way that facilitates an efficient computation.
We consider two possible formulations to define $v(S)$ for our purpose, namely $v_1$ in \eqref{eq:vfunc1} and $v_2$ in \eqref{eq:vfunc2}, and elaborate the latter as we find it more convenient.

\subsection{Notation}

Let us introduce notations used hereafter.
Without loss of generality, we consider that a data-point $\bm{x} \in \mathbb{R}^d$ is a vertical concatenation of $\bm{x}_{S^c} \in \mathbb{R}^{d - \vert S \vert}$ and $\bm{x}_S \in \mathbb{R}^{\vert S \vert}$, i.e.,
\begin{equation}
    \bm{x} = \begin{bmatrix} \bm{x}_{S^c} \\ \bm{x}_S \end{bmatrix},
\label{eq:split_x}
\end{equation}
where $\bm{x}_S$ denotes the subvector of $\bm{x}$ corresponding to the indices in $S$ (e.g., $\bm{x}_{\{1,3\}}=[x_1 \ x_3]^\tr$), and $S^c = D \backslash S$ denotes the complement of S.
Analogously, we can rearrange and partition the matrices $\bm{C}$ and $\bm{B}$ in \eqref{eq:Cmat} and \eqref{eq:Bmat}, respectively, as
\begin{equation}\begin{aligned}
    \bm{C} &= \begin{bmatrix} \bm{C}_{S^c} & \bm{C}_{S^c, S} \\ \bm{C}_{S^c, S}^\tr & \bm{C}_S \end{bmatrix}
    \quad \text{and} \\
    \bm{B} &= \begin{bmatrix} \bm{B}_{S^c} & \bm{B}_{S^c, S} \\ \bm{B}_{S^c, S}^\tr & \bm{B}_S \end{bmatrix},
\end{aligned}\label{eq:split_CB}\end{equation}
where $\bm{C}_{S^c} \in \mathbb{R}^{(d - \vert S \vert) \times (d - \vert S \vert)}$, $\bm{C}_{S^c, S} \in \mathbb{R}^{(d - \vert S \vert) \times \vert S \vert}$, $\bm{C}_S \in \mathbb{R}^{\vert S \vert \times \vert S \vert}$, and analogously for $\bm{B}$.
Here, $\bm{C}_S$ is the submatrix of $\bm{C}$ corresponding to the features in $S$, and so on.

\subsection{Value Function by Retraining}

Let $\bm{W}^{(S)} \in \mathbb{R}^{\vert S \vert \times p}$ be the parameter of probabilistic PCA in \eqref{eq:ppca1} that was learned on data comprising $\bm{x}_S$.
Then, we may define a value function as
\begin{equation}\begin{gathered}
    v_1(S) = \frac1{\vert S \vert} e^{(S)}(\bm{x}_S),
    \\
    e^{(S)}(\bm{x}_S) = \trace \big( (\bm{I}-\bm{B}^{(S)}) \bm{x}_S \bm{x}_S^\tr \big),
\end{gathered}\label{eq:vfunc1}\end{equation}
where $\bm{B}^{(S)} = \bm{W}^{(S)} ((\bm{W}^{(S)})^\tr \bm{W}^{(S)})^{-1} (\bm{W}^{(S)})^\tr$.
In words, \eqref{eq:vfunc1} defines a value function $v_1(S)$ as the reconstruction error of $\bm{x}_S$ with regard to the PCA trained only on the subset of features, $S$.
Obviously, we need to retrain PCA for each $S$ to compute the Shapley value of \eqref{eq:vfunc1}, which causes difficulties as follows:
\begin{itemize}
    \item As training PCA needs eigendecomposition of a $\vert S \vert \times \vert S \vert$ matrix or iterations of the EM algorithm, retraining may become inefficient for a moderate $d$.
    \item We have to maintain many models or training data itself, which is inefficient in terms of memory.
    \item It is challenging to consider a definite strategy to determine the dimensionality of the latent variable of PCA for different $S$, especially for small $\vert S \vert$.
\end{itemize}

\subsection{Value Function by Marginalization}

\begin{algorithm}[t]
    \caption{Monte Carlo approximation of Shapley values of reconstruction errors of PCA}
    \label{alg:main}
    \begin{algorithmic}[1]
        \renewcommand{\algorithmicrequire}{\textbf{Input:}}
        \renewcommand{\algorithmicensure}{\textbf{Output:}}
        \REQUIRE estimated parameters of PCA ($\bm{W}\in\mathbb{R}^{d \times p}$, $\sigma^2\in\mathbb{R}_{>0}$; we assume $\bm{b}=\bm{0}$), a test data-point $\bm{x}$, a target feature index $i$, and the number of Monte Carlo iterations $Q$
        \ENSURE a Shapley value $\varphi_i(v_2)$
        \STATE $\bm{C} \leftarrow \sigma^2\bm{I}+\bm{W}\bm{W}^\tr$
        \STATE $\bm{B} \leftarrow \bm{W}(\bm{W}^\tr\bm{W})^{-1}\bm{W}^\tr$
        \STATE $\varphi_i \leftarrow 0$
        \FOR {$q = 1$ to $Q$}
            \STATE draw an order $O_q$ randomly from $\pi(D)$
            \STATE $\varphi_{i,q} \leftarrow v ( \text{Pre}_i(O_q) \cup \{i\} ) - v ( \text{Pre}_i(O_q) )$, using \eqref{eq:vfunc2_comp}
            \STATE $\varphi_i \leftarrow \varphi_i + \varphi_{i,q}/Q$
        \ENDFOR
    \end{algorithmic}
\end{algorithm}

\begin{figure*}[t]
    \centering
    \begin{minipage}[t]{0.27\textwidth}
        \centering\vspace*{0pt}
        \includegraphics[clip,height=2.8cm]{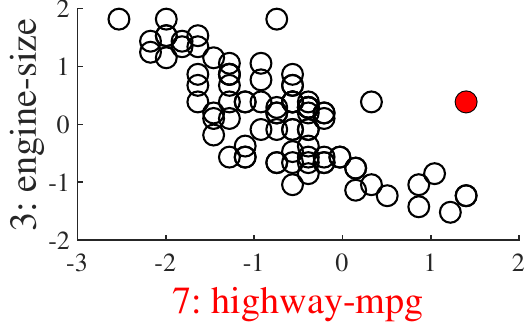}
        \subcaption{}
    \end{minipage}
    \hspace{0.01\textwidth}
    \begin{minipage}[t]{0.305\textwidth}
        \centering\vspace*{0pt}
        \includegraphics[clip,height=2.8cm]{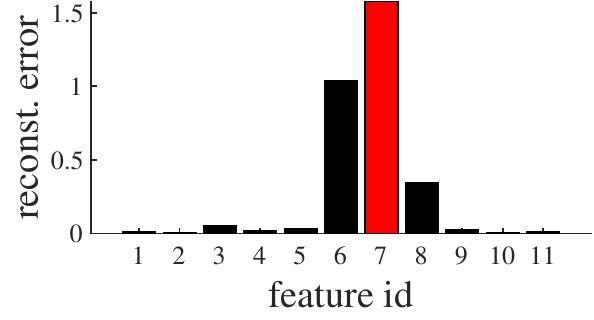}
        \subcaption{}
    \end{minipage}
    \hspace{0.01\textwidth}
    \begin{minipage}[t]{0.305\textwidth}
        \centering\vspace*{0pt}
        \includegraphics[clip,height=2.8cm]{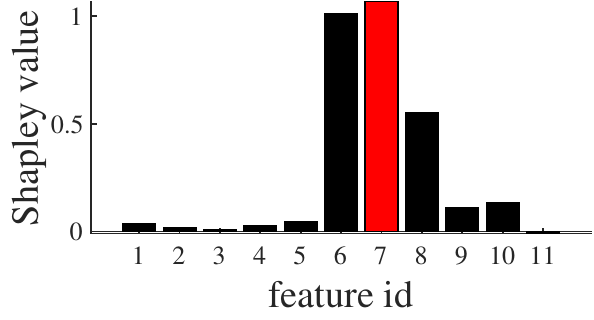}
        \subcaption{}
    \end{minipage}
    \caption{(a) Scatter plot of normalized test data in a test trial, wherein a filled circle indicates the artificial anomalous data-point. In this test trial, the feature \#7 of the indicated data-point was made anomalous. (b) Reconstruction errors of each feature of the anomalous data-point. (c) Shapley values of reconstruction error of each feature of the anomalous data-point.}
    \label{fig:car_1_7}
\end{figure*}
\begin{figure*}[t]
    \centering
    \begin{minipage}[t]{0.27\textwidth}
        \centering\vspace*{0pt}
        \includegraphics[clip,height=2.8cm]{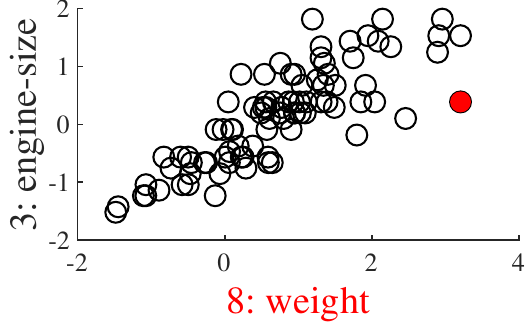}
        \subcaption{}
    \end{minipage}
    \hspace{0.01\textwidth}
    \begin{minipage}[t]{0.305\textwidth}
        \centering\vspace*{0pt}
        \includegraphics[clip,height=2.8cm]{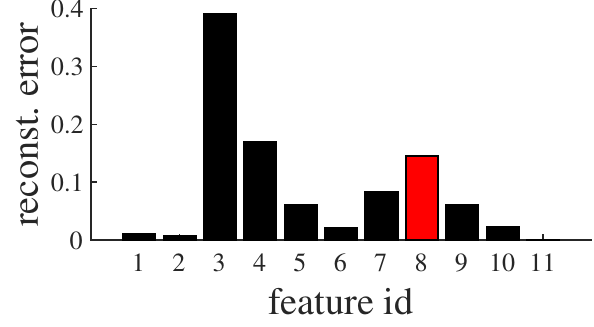}
        \subcaption{}
    \end{minipage}
    \hspace{0.01\textwidth}
    \begin{minipage}[t]{0.305\textwidth}
        \centering\vspace*{0pt}
        \includegraphics[clip,height=2.8cm]{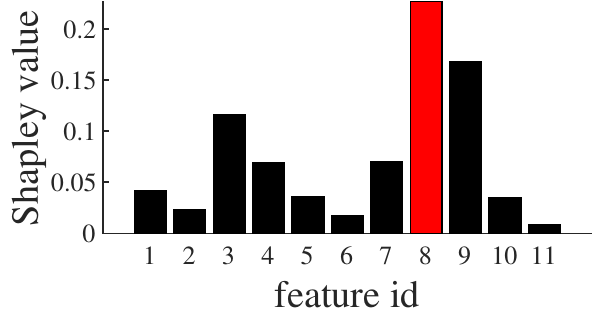}
        \subcaption{}
    \end{minipage}
    \caption{As in \figurename~\ref{fig:car_1_7}, but for another test trial, in which the feature \#8 of the indicated data-point was made anomalous.}
    \label{fig:car_1_8}
\end{figure*}
\begin{figure*}[t]
    \centering
    \begin{minipage}[t]{0.27\textwidth}
        \centering\vspace*{0pt}
        \includegraphics[clip,height=2.8cm]{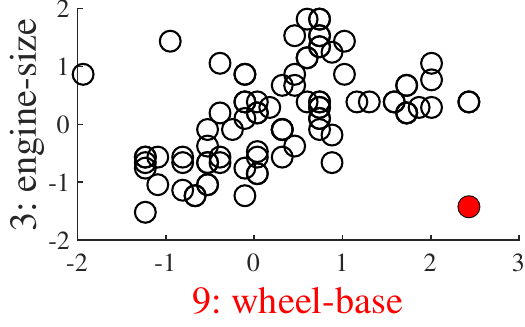}
        \subcaption{}
    \end{minipage}
    \hspace{0.01\textwidth}
    \begin{minipage}[t]{0.305\textwidth}
        \centering\vspace*{0pt}
        \includegraphics[clip,height=2.8cm]{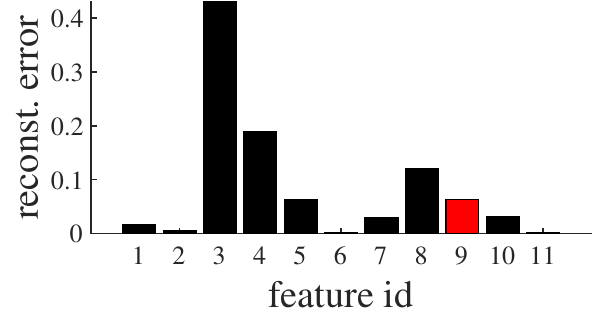}
        \subcaption{}
    \end{minipage}
    \hspace{0.01\textwidth}
    \begin{minipage}[t]{0.305\textwidth}
        \centering\vspace*{0pt}
        \includegraphics[clip,height=2.8cm]{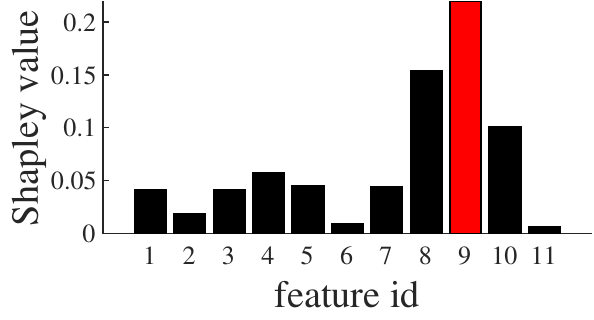}
        \subcaption{}
    \end{minipage}
    \caption{As in \figurename~\ref{fig:car_1_7}, but for yet another test trial, in which the feature \#9 of the indicated data-point was made anomalous.}
    \label{fig:car_48_9}
\end{figure*}

To alleviate the difficulties of the retraining strategy, we consider to avoid retraining by performing marginalization with regard to unused features, as done in literature \citep{strumbelj_efficient_2010,strumbelj_explaining_2014,lundberg_unified_2017}.
That is, we define a value function as
\begin{equation}
    v_2(S) = \frac1d \mathbb{E}_{p(\bm{x}_{S^c} \mid \bm{x}_S)} [ e(\bm{x}) ].
\label{eq:vfunc2}
\end{equation}
We note that $e(\bm{x})$ is simply the reconstruction error on the full test data-point, $\bm{x}$, computed with the PCA trained on all the features, as defined in \eqref{eq:recerr_general}.

In literature \citep{strumbelj_efficient_2010,strumbelj_explaining_2014,lundberg_unified_2017}, the marginalization relies on an assumption of statistical independence, $p(\bm{x}_{S^c} \mid \bm{x}_S) = p(\bm{x}_{S^c})$.
However, when an anomaly detection based on reconstruction error is valid, there usually exists dependence between features, which may make the approaches based on the independence assumption \citep{strumbelj_efficient_2010,strumbelj_explaining_2014,lundberg_unified_2017} less meaningful.

Meanwhile, we can exactly compute \eqref{eq:vfunc2} \emph{without the independence assumption} using the probabilistic view of PCA in \eqref{eq:ppca1}--\eqref{eq:ppca3} as follows.
First, the conditional distribution of $\bm{x}_{S^c}$ given $\bm{x}_S$, namely $p(\bm{x}_{S^c} \mid \bm{x}_S)$, is immediately derived from \eqref{eq:ppca3} as
\begin{equation}
    p(\bm{x}_{S^c} \mid \bm{x}_S) = \mathcal{N}_{\bm{x}_{S^c}} ( \bm{C}_{S^c, S} \bm{C}_S^{-1} \bm{x}_S, \ \bm{C}_{S^c} - \bm{C}_{S^c, S} \bm{C}_S^{-1} \bm{C}_{S^c, S}^\tr ),
\label{eq:ppca4}
\end{equation}
where we assume that the data are centered so that $\bm{b}=\bm{0}$ without loss of generality.
Note that $\bm{C}_{\cdot}$'s are the submatrices of $\bm{C}$ defined in \eqref{eq:split_CB}.
Second, the total reconstruction error, $e(\bm{x})$ in \eqref{eq:recerr_general}, can be decomposed as
\begin{equation}\begin{aligned}
    e(\bm{x}) &= \trace \big( (\bm{I} - \bm{B}) \bm{x} \bm{x}^\tr \big)
    \\
    &= \trace \big( (\bm{I} - \bm{B}_{S^c}) \bm{x}_{S^c} \bm{x}_{S^c}^\tr \big)
    \\ &\quad - 2 \trace ( \bm{B}_{S^c, S} \bm{x}_S \bm{x}_{S^c} )
    + \trace \big( (\bm{I} - \bm{B}_S) \bm{x}_S \bm{x}_S^\tr \big),
\end{aligned}\label{eq:split_e}\end{equation}
where $\bm{B}_{\cdot}$'s are the submatrices of $\bm{B}$ in \eqref{eq:split_CB}.
Finally, using \eqref{eq:ppca4} and \eqref{eq:split_e}, $v_2(S)$ in \eqref{eq:vfunc2} is computed by
\begin{equation}\begin{aligned}
    &v_2(S) = \frac1d \mathbb{E}_{p(\bm{x}_{S^c} \mid \bm{x}_S)} [ e(\bm{x}) ]
    \\
    &= \frac1d \Big(
        \trace \big( (\bm{I} - \bm{B}_{S^c}) \bm{V}_{S^c} \big)
        + \trace \big( (\bm{I} - \bm{B}_{S^c}) \bm{m}_{S^c} \bm{m}_{S^c}^\tr \big)
        \\ &\qquad
        - 2 \trace ( \bm{B}_{S^c, S} \bm{x}_S \bm{m}_{S^c}^\tr )
        + \trace \big( (\bm{I} - \bm{B}_S) \bm{x}_S \bm{x}_S^\tr \big)
    \Big),
\end{aligned}\label{eq:vfunc2_comp}\end{equation}
where we use notations
\begin{align}
    \bm{m}_{S^c} &= \bm{C}_{S^c, S} \bm{C}_S^{-1} \bm{x}_S,
    \\
    \bm{V}_{S^c} &= \bm{C}_{S^c} - \bm{C}_{S^c, S} \bm{C}_S^{-1} \bm{C}_{S^c, S}^\tr.
\end{align}

Unlike the retraining strategy, the marginalization strategy does not require to keep many models or training data because \eqref{eq:vfunc2_comp} can be computed only from the parameters of the full model.
A computational bottleneck of \eqref{eq:vfunc2_comp} lies in the computation of $\bm{C}_{S^c, S} \bm{C}_S^{-1}$, but this can be computed relatively efficiently in general.

\subsection{Approximation of Shapley values}

Given the value function in \eqref{eq:vfunc2_comp}, we approximate its Shapley values using a Monte Carlo approximation of \eqref{eq:shapley2}.
We summarize the overall procedures in Algorithm~\ref{alg:main}.
For the detailed deviations of the Monte Carlo approximation, consult e.g., \citet{strumbelj_efficient_2010,strumbelj_efficient_2010}.


\section{Numerical Examples}
\label{examples}

\subsection{Artificial Anomalies}

We concretely investigated how the Shapley values were different from the reconstruction errors using a multivariate dataset to which we inserted artificial anomalies.
We used the 2004 New Car and Truck Data, which is available online\footnote{\url{jse.amstat.org/jse_data_archive.htm}}, as the base dataset.
It contains specifications of vehicles released in 2004, such as price, fuel efficiency, engine size, and car size; in total, it comprises $d=11$ features.
The original dataset consists of 428 instances.
We eliminated 41 instances including missing values and used the first 300 instances for the training and the last 87 for the test.
We repeated the test phase for different settings of how an artificial anomaly is created; in each test, we replaced the value of a feature of a data-point to the maximum value of the same feature in the test set.
Thus, we conducted $11 \times 87= 957$ test trials in total.
The dimensionality of PCA was set to $p=8$.

In Figs.~\ref{fig:car_1_7}, \ref{fig:car_1_8}, and \ref{fig:car_48_9}, we exemplify three test trials.
Each figure comprises: (a) a scatter plot of an ``anomalous'' feature against the feature \#3 (engine size), (b) the reconstruction errors of each feature on the anomalous instance, and (c) the corresponding Shapley values computed via the proposed method.
In Fig.~\ref{fig:car_1_7}, an explanation that would be derived from the reconstruction errors and the one from the Shapley values roughly coincide, i.e., the feature \#7 (highway miles per gallon), which was made anomalous in fact, indicates the largest value in both criteria.
In contrast, in Figs.~\ref{fig:car_1_8} and \ref{fig:car_48_9}, the reconstruction errors are suspicious whereas the Shapley values are somewhat trustworthy; e.g., in Fig.~\ref{fig:car_1_8}, though the anomaly was inserted to the feature \#8 (weight), the reconstruction error shows its largest value at feature \#3.



We also examined how often anomalous features are identified correctly by looking at the reconstruction errors or the Shapely values.
In Table~\ref{tab:hits}, we report the Hits@$n$, which denotes the rate that a truly anomalous feature exists in the set of features having the $n$ largest values of a criterion (the errors or the Shapley).
Table~\ref{tab:hits} is for two experimental cases, \textsc{Max} and \textsc{Min}: \textsc{Max} denotes the experimental setting explained above, i.e., an anomaly is generated by replacing a feature value by its maximum.
\textsc{Min} is its slightly differed version, in which an anomaly is generated by the replacement with minimum values.
From Table~\ref{tab:hits}, we can observe that the Shapley values are advantageous to identify anomalous features.

\subsection{Anomaly Detection Datasets}

We also applied the proposed method to real-world anomaly detection datasets.
We downloaded the datasets summarized as Outlier Detection DataSets \citep{ODDS}, which are available online\footnote{\url{odds.cs.stonybrook.edu/}}.
From the category of multi-dimensional point datasets, we selected 10 datasets with which a PCA-based anomaly detection worked well.
Each dataset comprises a set of $m_\text{good}$ normal instances and a set of $m_\text{bad}$ anomalous instances.
We used the first $m_\text{good} - m_\text{bad}$ elements of the normal set for training, and the remaining elements of the normal set and the whole anomalous set for a test.
The dimensionality of PCA was chosen so that it could explain about 95\% of the variance of the training data.

As conducting case studies for each dataset is unrealistic due to the lack of precise understanding of causes of the anomalies, we just examined the discrepancy between reconstruction errors and their Shapley values in test sets.
Concretely, we investigated the correlation coefficient between them.
In Table~\ref{tab:corr}, we show the median values\footnote{We computed the medians of the correlation coefficients via the Fisher transformation $z=\log((1+r)/(1-r))/2$ and the inverse transformation $r_\text{median}=\operatorname{tanh}(z_\text{median})$.} of the correlation coefficients computed for each feature individually.
There, the correlation on a full test set ($r_\text{all}$), the one on the normal part of the test set ($r_\text{good}$), and the one on the anomalous part of the test set ($r_\text{bad}$) are reported.
We can observe that the reconstruction errors and their Shapely values are strongly correlated (say, with $r>0.9$) in many cases, whereas the correlation is moderate (say, $r<0.7$) in some cases.
These results do not completely preclude the use of reconstruction errors for explaining anomalies, but one should be aware that there are cases where the Shapley values do not completely agree with the reconstruction errors of each feature.


\begin{table}[t]
    \centering
    \caption{Hits@$n$ for two experimental cases, \textsc{Max} and \textsc{Min}.}
    \label{tab:hits}
    \begin{tabular}{ccccc}
        \toprule
        & \multicolumn{2}{c}{\textsc{Max}} & \multicolumn{2}{c}{\textsc{Min}} \\
        \cmidrule(lr){2-3}
        \cmidrule(lr){4-5}
        & Hits@1 & Hits@3 & Hits@1 & Hits@3 \\
        \midrule
        reconstruction error & .316 & .605 & .271 & .471 \\
        Shapley values & .484 & .801 & .484 & .710 \\
        \bottomrule
    \end{tabular}
\end{table}

\begin{table}[t]
    \centering
    \vspace*{2ex}
    \caption{Datasets property ($d$ and $m=m_\text{good}+m_\text{bad}$) and the correlation coefficients between reconstruction errors and their Shapley values. The medians (via the Fisher transformation) of the coefficients computed for each feature are reported. $r_\text{all}$, $r_\text{good}$, and $r_\text{bad}$ denote the correlation coefficients for the whole, the normal part, and the anomalous part of the test sets, respectively.}
    \label{tab:corr}
    \begin{tabular}{cccccc}
        \toprule
        dataset & $d$ & $m$ & $r_\text{all}$ & $r_\text{good}$ & $r_\text{bad}$ \\
        \midrule
        \textsc{Cardio} & 21 & 1831 & .866 & .893 & .797 \\
        \textsc{ForestCover} & 10 & 286048 & .756 & .536 & .808 \\
        \textsc{Ionosphere} & 33 & 351 & .984 & .986 & .985 \\
        \textsc{Mammography} & 6 & 11183 & .854 & .268 & .854 \\
        \textsc{Musk} & 166 & 3062 & .945 & .987 & .949 \\
        \textsc{Satimage-2} & 36 & 5803 & .975 & .993 & .981 \\
        \textsc{Shuttle} & 9 & 49097 & .869 & .958 & .893 \\
        \textsc{Vowels} & 12 & 1456 & .883 & .833 & .877 \\
        \textsc{WBC} & 30 & 278 & .956 & .955 & .943 \\
        \textsc{Wine} & 13 & 129 & .817 & .785 & .657 \\
        \bottomrule
    \end{tabular}
\end{table}



\section{Discussion}
\label{discussion}

\subsection{Related Work}
\label{discussion:related}


Several researchers have been working on explaining anomaly detection.
\citet{dang_local_2013}, \citet{micenkova_explaining_2013}, and \citet{dang_discriminative_2014} proposed ways to explain anomaly detection based on nearest neighbors by looking for a set of features (i.e., a subspace) that maximally discriminates an outlier from neighboring inliers.
Similarly, \citet{liu_contextual_2018} developed a method to interpret anomalies by learning classifiers that distinguish normal and anomalous data-points.
Moreover, \citet{macha_explaining_2018} proposed a framework to generate descriptions of outliers via clustering in subspaces and its summarization.
In another context, \citet{tang_mining_2015} proposed an interpretable definition of outliers for categorical data, which is understandable in terms of common and uncommon components of anomalous and normal values.
Furthermore, \citet{gupta_beyond_2019} developed a method to optimize pictorial explanations of anomalies.
The types of anomaly detection methods premised in these studies are basically different from ours (i.e., reconstruction error-based), but seeking connections is an interesting future direction of research.

One of the existing studies that are mostly related to ours is the one by \citet{antwarg_explaining_2019}, in which they use kernel SHAP \citep{lundberg_unified_2017} to explain the reconstruction errors of an autoencoder.
A possible difficulty of their methods lies in the assumption of feature independence premised in kernel SHAP.
As features are often dependent in anomaly detection applications, considering feature correlations may be essential in some practices.
We will be working on comparing our current method (and its extensions) with kernel SHAP.

\subsection{Direct and Indirect Extensions}

One of the promising extensions of our method is application to variants of PCA.
For example, the mixtures of PCAs \citep{tipping_mixtures_1999} are useful to capture multimodality and in fact, utilized in anomaly detection applications (see, e.g., \citet{yairi_data-driven_2017}).
An extension to sparse PCA \citep{guan_sparse_2009} is also interesting to ensure interpretability of both a model and reconstruction errors.

Extensions to nonlinear and/or non-Gaussian models, with which conditional distributions cannot be obtained easily, are also an interesting and demanding direction of research.
To this end, one possible option is to approximate the conditional distributions in some tractable ways.
For example, approximation for using robust distributions (e.g., $t$-distribution) would be useful to improve anomaly detection performance.

Reconstruction errors are always nonnegative, but the raw Shapley values may take negative values, which is basically challenging to interpret.
Also, explanations by the Shapley values are usually not parsimonious.
To overcome these issues, approximation methods such as kernel SHAP and their variants would be useful.


\section*{Acknowledgment}

This work was supported by JSPS KAKENHI Grant Number JP19K21550.

\printbibliography


\end{document}